\documentclass[runningheads]{llncs}

\usepackage[T1]{fontenc}
\usepackage{graphicx}
\usepackage{color}
\usepackage[colorlinks,urlcolor=blue,citecolor=blue,linkcolor=blue]{hyperref}

\usepackage{booktabs}
\usepackage{array}
\usepackage{tabularx}
\usepackage{makecell}
\usepackage{amsmath,amssymb}

\begin{document}

\title{TLoRA: Tri-Matrix Low-Rank Adaptation of Large Language Models}

\author{Tanvir Islam}
\institute{Okta, Bellevue, WA, USA\\
\email{tanvir.islam@okta.com}}
\maketitle

\renewcommand{\thefootnote}{} 
  \footnotetext{Proceedings of the Pacific Rim International Conference on Artificial Intelligence (PRICAI 2025), Wellington, New Zealand.}
\renewcommand{\thefootnote}{\arabic{footnote}}

\begin{abstract}
We propose \textbf{TLoRA}, a novel tri-matrix low-rank adaptation method
that decomposes weight updates into three matrices: two fixed random
matrices and one trainable matrix, combined with a learnable, layer-wise
scaling factor. This tri-matrix design enables TLoRA to achieve highly
efficient parameter adaptation while introducing minimal additional
computational overhead. Through extensive experiments on the GLUE
benchmark, we demonstrate that TLoRA achieves comparable performance to
existing low-rank methods such as LoRA and adapter-based techniques,
while requiring significantly fewer trainable parameters. Analyzing the
adaptation dynamics, we observe that TLoRA exhibits Gaussian-like weight
distributions, stable parameter norms, and scaling factor variability
across layers, further highlighting its expressive power and
adaptability. Additionally, we show that TLoRA closely resembles LoRA in
its eigenvalue distributions, parameter norms, and cosine similarity of
updates, underscoring its ability to effectively approximate
LoRA's adaptation behavior. Our results establish TLoRA
as a highly efficient and effective fine-tuning method for LLMs,
offering a significant step forward in resource-efficient model
adaptation.


\end{abstract}

\section{Introduction}

Fine-tuning is a critical process in the adaptation of large language
models, aiming to tailor the model to perform specific tasks or solve
specific problems \cite{naveed_comprehensive_2023,zhang_instruction_2023}. The technique involves adjusting the
pre-trained model’s weights by exposing it to
task-specific data, thereby refining its understanding and responses
based on the given context. This method leverages the foundational
knowledge encoded within the model while introducing new patterns and
nuances pertinent to the targeted problem statement. Foundational
models, such as GPT \cite{brown_language_2020}, BERT \cite{devlin2019bert},
LLaMA \cite{touvron_llama_2023}, and RoBERTa \cite{liu_roberta_2019}, have been
pre-trained on vast corpora and capture a rich representation of
language, including syntax, semantics, and general world knowledge.
These models act as a robust starting point, offering a versatile
understanding of language that can be fine-tuned for various downstream
tasks, including natural language understanding and generation.

In recent years, there has been significant interest among practitioners
and researchers in exploring various fine-tuning methods for large
language models (LLMs) \cite{naveed_comprehensive_2023,wei_empirical_2023}. One common
approach is full fine-tuning, which involves continued training of the
model to specialize it for a specific task, such as sentiment analysis
\cite{prottasha_transfer_2022} or question answering \cite{luo_chatkbqa_2023}, by
using task-specific data. This method, while effective, can be
computationally intensive for LLMs. By contrast, Parameter-Efficient
Fine-Tuning (PEFT) presents an alternative cost-effective solution. In
fact, PEFT methods are found to be better than in-context learning \cite{liu_few-shot_2022}. In the PEFT, instead of updating all the model parameters,
only a subset or a small set of additional parameters is adjusted. This selective
fine-tuning reduces the computational overhead while still achieving
high performance, making it an attractive option for resource and
time-constrained environments.

Low-Rank Adaptation, or LoRA \cite{hu_lora_2021} is one such method that has significantly
advanced the field of Parameter-Efficient Fine-Tuning. In LoRA, certain layers of
the model, typically within dense or attention layers, are re-parameterized
by introducing low-rank matrices that effectively reduce the
dimensionality of the weight updates needed for fine-tuning. Instead of
updating the full set of parameters, LoRA learns a low-rank decomposition
of parameter updates, which minimizes the number of trainable parameters
while still allowing the model to capture task-specific nuances. Variants of LoRA have also been reported in the literature \cite{lialin_relora_2023,kopiczko_vera_2023,li_vb-lora_2024}.

Inspired by the promising results achieved by LoRA \cite{hu_lora_2021} we
ask: Can we do even better? Can we further reduce the number of
trainable parameters while maintaining the similar performance? This
paper introduces TLoRA: Tri-Matrix Low-Rank Adaptation of Large Language
Models, a novel technique that leverages a tri-matrix structure along
with a clever adaptive scaling mechanism. TLoRA, introduced in this
paper, offers a more flexible and efficient approach for adapting large
language models to diverse tasks. By exploring this innovative method, we aim to push the
boundaries of efficient model fine-tuning, delivering superior
performance with reduced resource requirements.

\section{TLoRA}
\subsection{Objective functions}
We know that a pre-trained language model
\(P_{\theta}\left( y \middle| \, x \right)\) can be adapted for various
downstream tasks by fine-tuning it on task-specific data. In these
tasks, we generally use a training dataset of context–target pairs:
\[
  Z = \{(x_i, y_i)\}_{i=1,\dots,N},
\]
where \(x_i\) consists of a sequence of tokens and \(y_i\) could 
either be a sequence of tokens corresponding to the answer generated by
the model or a classification target, depending on the task. For
example, in a natural language inference (NLI) task \cite{nangia_human_2019}, \(x_{i}\) represents a pair of
sentences (e.g., a premise and a hypothesis), and \(y_{i}\) is the
corresponding classification label. The LLM model is then updated to
optimize its performance for this specific task.

If we opt for full fine-tuning, the pre-trained model parameters
\(\theta_{0}\) are updated directly to \(\theta_{0} + \Delta\theta\) by
maximizing the conditional language modeling objective. In the context
of classification tasks, such as those in the GLUE benchmark, we seek to
optimize the model to predict a discrete class label \(y\) based on an
input sequence \(x\). Therefore, the objective becomes:

\[\max_{\theta}{\sum_{(x,y) \in Z}^{}{\log P_{\theta}}}\left( y \middle| \, x \right)\]

where \(P_{\theta}\left( y \middle| \, x \right)\) is the probability
distribution over class labels given the input \(x\), and \(\theta\)
represents the full set of model parameters, including the pre-trained
ones \(\theta_{0}\) and the task-specific updates \(\Delta\theta\).

However, this full fine-tuning approach is computationally expensive,
especially for large models with billions of parameters like GPT-3 \cite{brown_language_2020} and LLaMA \cite{touvron_llama_2023}. To
address this issue, low-rank adaptation technique is a more efficient
alternative where only a small set of task-specific parameters \(\phi\)
is learned. Instead of updating \(\theta_{0}\) directly, TLoRA uses a low‐rank update \(\Delta\theta(\phi)\), which is much smaller than the original parameter set \(\theta_{0}\). The optimization problem then becomes:

\[
  \max_{\phi}
    \sum_{(x,y)\in Z}
      \log P_{\theta_{0} + \Delta\theta(\phi)}\bigl(y \mid x\bigr)
\]

where \(\Delta\theta(\phi)\) is the task‐specific parameter update encoded by \(\phi\), and the term
\(P_{\theta_{0} + \Delta\theta(\phi)}\bigl(y \mid x\bigr)\) is the probability distribution over class labels for the adapted model.

Like the LoRA, in TLoRA, the update \(\Delta\theta(\phi)\) is represented in a low-rank form to make the adaptation both memory and
computation-efficient. By restricting the update to a low-rank
representation, we ensure that the number of learnable parameters
\textbar{}\(\phi|\) is much smaller than the size of the original model
\(\left| \theta_{0} \right|\). In this setting, the model's
classification objective remains the same as in full fine-tuning, but
now, instead of optimizing all of \(\theta\), we only need to optimize
the much smaller parameter set \(\phi\), which encodes the low-rank
adaptation.

\subsection{Tri-matrix decomposition and adaptive scaling}
TLoRA is an extension of the Low-Rank Adaptation (LoRA) technique \cite{hu_lora_2021} that uses a tri-matrix decomposition for more flexible and
efficient adaptation of pre-trained language models to downstream tasks.
TLoRA allows for a more granular adaptation of the
model\textquotesingle s weights while maintaining a minimal increase in
parameters by introducing three low-rank matrices. Additionally, TLoRA
incorporates a trainable scaling factor to control the magnitude of the
low-rank update. Figure 1 provides a visual representation of
TLoRA\textquotesingle s fine-tuning process.

\begin{figure}[h]
		\includegraphics[width=0.5\linewidth]{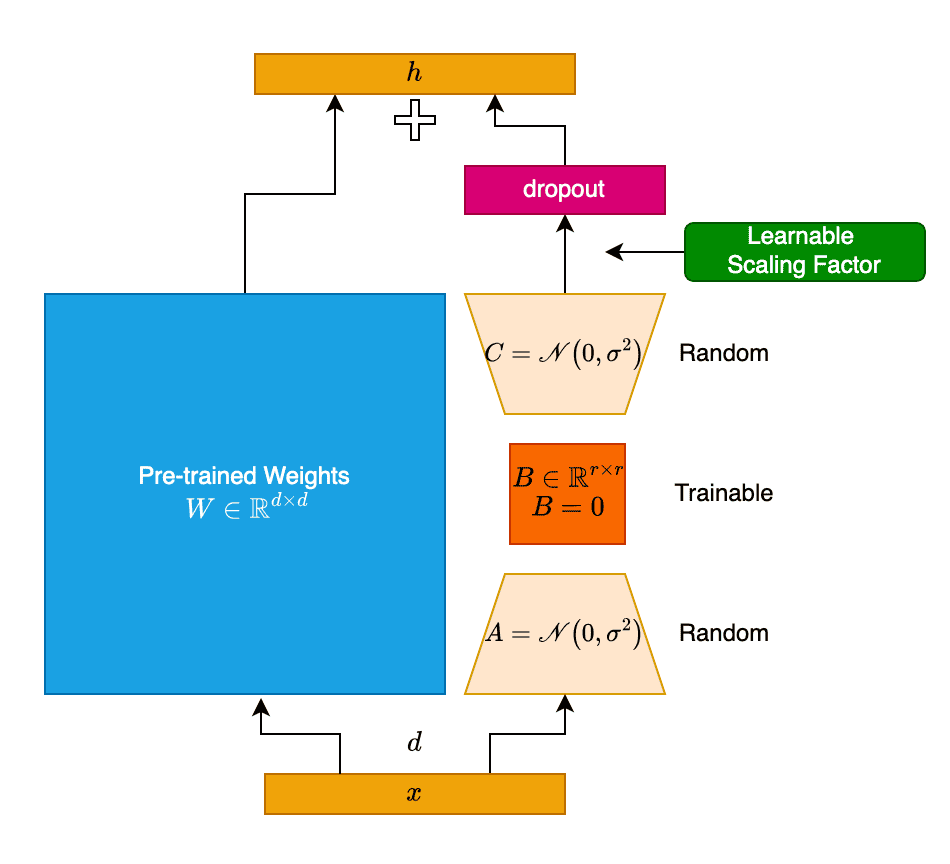}
    \centering
		\caption{Schematic representation of TLoRA. The weight update
is decomposed into a tri-matrix structure consisting of two fixed random
matrices \(A\) and \(C\), and a trainable matrix B. The input \(x\) is
projected through the sequence of matrices \(A,\ B,\ C\), followed by a
learnable scaling factor to control the magnitude of the adaptation.
This tri-matrix design enables efficient low-rank adaptation while
minimizing trainable parameters.}
		\label{fig1}
\end{figure}

\subsubsection{TLoRA Decomposition and Weight Update}

Given a pre-trained language model with weight matrix
\(W_{0} \in R^{d \times k}\), TLoRA computes the weight update
\(\Delta W\) through a tri-matrix decomposition. Specifically, the
update is decomposed into three low-rank matrices:

\[A \in R^{d \times r},\quad B \in R^{r \times r},\quad C \in R^{r \times k}\]

where \(r\) is the rank of the decomposition, typically much smaller
than \(d\), (\(\ r\  \ll d\)) to ensure that the number of parameters
added during adaptation remains small. The original weight matrix
\(W_{0}\) of the layer remains frozen.

In TLoRA, only the matrix \(B\) is trainable, while matrices \(A\) and
\(C\) are randomly initialized and fixed (non-trainable) throughout
adaptation. This design leverages the efficiency of low-rank updates
without requiring significant parameter growth. The technical rationale
for keeping \(A\) and \(C\) fixed and random is to create a structured
yet efficient transformation that enhances model flexibility with
minimal additional parameters. By making \(B\) trainable, TLoRA allows
the model to learn a task-specific transformation within a constrained
low-rank space defined by \(A\) and \(C\).

The low-rank update is computed as the product of these three matrices:

\[\Delta W\  = \ A\ B\ C\]

The adapted weight matrix \(W_{\mathrm{adapted}}\) is then the sum of the
original weight matrix \(W_{0}\) and the low-rank update \(\Delta W\):

\[W_{adapted} = W_{0} + \alpha\Delta\, W = W_{0} + \alpha\, ABC\]

where \(\alpha\) is a trainable scaling factor that controls the
contribution of the low-rank adaptation.

In the forward pass of the TLoRA layer, the model performs the following
steps. The input \(x \in R^{k}\) is passed through the original
pre-trained linear transformation represented by \(W_{0}\):

\[h_{0} = W_{0}x\]

Here, \(x\) is the input (e.g., a token embedding), and
\(h_{0} \in R^{d}\) is the output of the standard linear transformation.

The low-rank adaptation is computed by multiplying the input \(x\) with
the tri-matrix decomposition:

\[\Delta h = \alpha  (ABC)  x\]

Dropout is then applied to the low-rank update to regularize the model:

\[
  \Delta h_{\mathrm{dropout}}
    = \mathrm{Dropout}\bigl(\Delta h\bigr)
\]

The final output of the TLoRA layer is the sum of the standard linear
transformation and the low-rank update with dropout:

\[
  h
    = h_{0} + \Delta h_{\mathrm{dropout}}
    = W_{0}x + \alpha\,A\,B\,C\,x
\]

\subsubsection{Trainable Scaling Factor}

The scaling factor \(\alpha\) in TLoRA is a trainable parameter, which
allows the model to dynamically adjust the contribution of the low-rank
adaptation during training. This means that unlike in the original LoRA
where \(\alpha\) is determined through a fixed hyperparameter, in TLoRA,
\(\alpha\) is learned during training, allowing the model to adjust the
strength of the low-rank update dynamically. This trainable scaling
factor enables the model to fine-tune the balance between preserving the
pre-trained knowledge and incorporating task-specific adaptations.
During training, \(\alpha\) is optimized through gradient descent along
with other model parameters. By learning the optimal scaling factor for
the target layer, TLoRA can flexibly control the strength of the
low-rank update, allowing for more precise adaptation to specific tasks
without introducing excessive complexity.

\begin{figure}[h]
  \centering
  \includegraphics[width=0.5\linewidth]{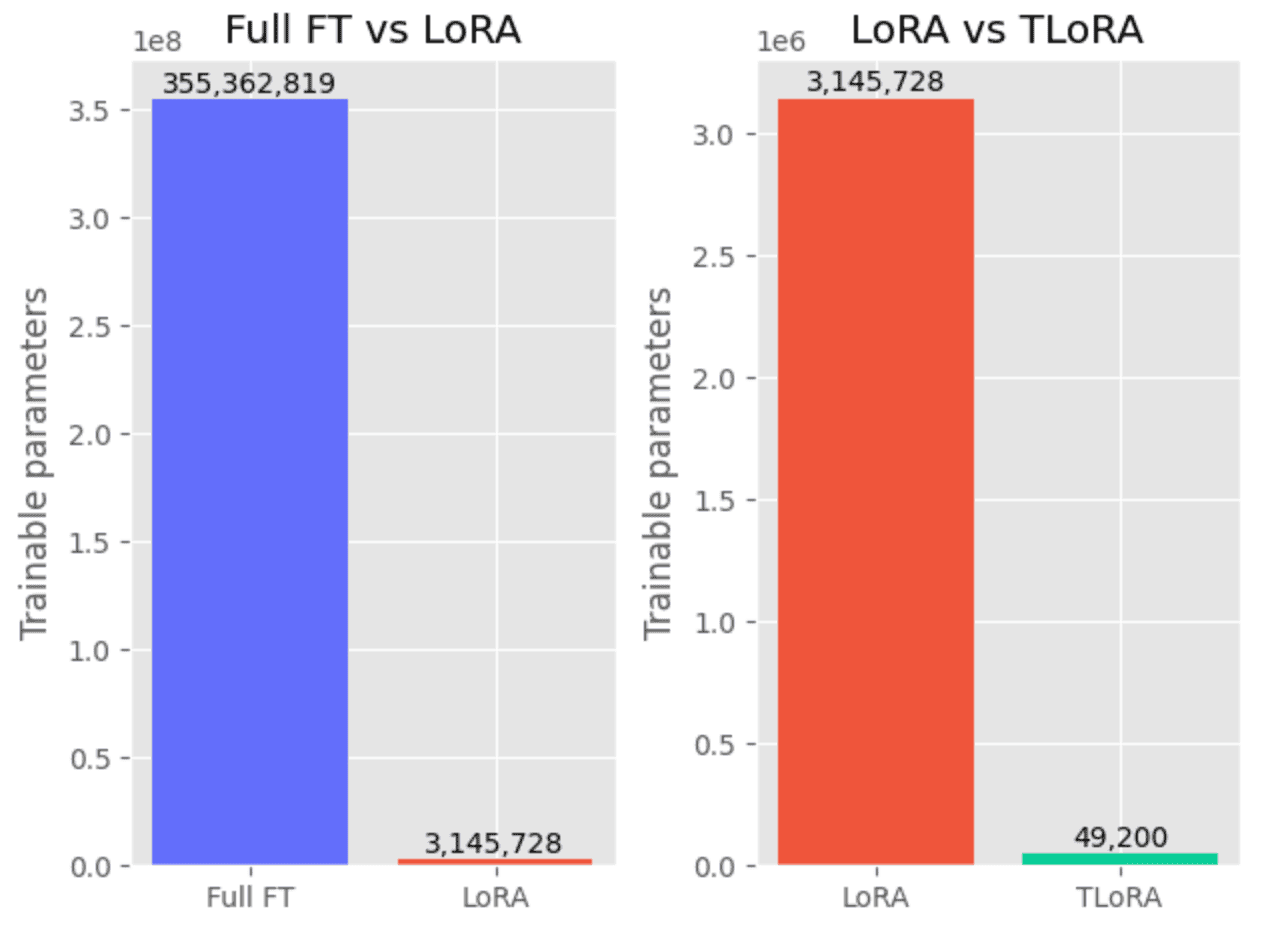}
  \caption{Comparison of trainable parameters across different
fine-tuning methods. The bar chart illustrates the parameter count for
full fine-tuning (Full FT), LoRA (rank \(r\)=32), and TLoRA (rank
\(r = 32\)). TLoRA significantly reduces the number of trainable
parameters compared to LoRA, showcasing its parameter efficiency while
maintaining competitive performance.}
  \label{fig:2}
\end{figure}

\subsection{Parameter count}
In TLoRA, we achieve significant parameter reduction compared to
conventional fine-tuning and LoRA while maintaining similar adaptation
capacity. For example, in a model like RoBERTa-large \cite{liu_roberta_2019} with
355 million parameters, a standard fine-tuning approach requires
training all parameters, resulting in a high computational burden and
storage cost. In contrast, LoRA achieves a reduction by introducing only
low-rank matrices of rank \(r\), where the trainable parameter count is
proportional to \(r\). For instance, with \(r\) = 8, LoRA requires
786,432 trainable parameters, and this count increases with the increase
in \(r\), reaching 3,145,728 trainable parameters for \(r\) = 32.

TLoRA further optimizes parameter efficiency by leveraging a tri-matrix
decomposition where only the middle matrix
is trainable, reducing the parameter count even more. This structure
results in TLoRA needing only 3,120 trainable parameters at \(r\) = 8,
representing a 252x reduction in trainable parameters compared to LoRA. As \(r\) increases, TLoRA maintains substantial parameter
savings over LoRA, with a 128x reduction at \(r\) = 16 and a 64x
reduction at \(r\) = 32. A visual illustration is shown in Figure 2 for
\(r\) = 32.

Notably, even at a higher effective rank (e.g., TLoRA with \(r\) = 32),
TLoRA achieves a parameter count improvement of 16x over LoRA at \(r\) =
8, demonstrating TLoRA\textquotesingle s ability to match LoRA's
performance with significantly fewer trainable parameters, enhancing
memory efficiency and enabling scalability for large language models on
parameter-constrained hardware. The detailed comparison of parameters
count of TLoRA to LoRA is shown in Table 1.

\begin{table}[ht]
  \small
  \centering
  \caption{Theoretical trainable parameters between LoRA and TLoRA}
  \label{tab:params}
  \begin{tabularx}{\linewidth}{XXXXX}
    \toprule
    Rank & Full FT & LoRA     & TLoRA   & Improvement\\
    \midrule
     8   & 355 M    & 786 k    & 3.1 k   & 252×  \\
    16   & 355 M    & 1.57 M   & 12.3 k  & 128×  \\
    32   & 355 M    & 3.15 M   & 49.2 k  & 64×   \\
    \bottomrule
  \end{tabularx}
\end{table}

\subsection{Initialization}
In TLoRA, non-Trainable Matrices \(A\) and \(C\) are initialized with a
Kaiming (He) normal initialization \cite{he_delving_2015}. The trainable matrix
\(B \in R^{r \times r}\) is initialized to zeros. This ensures that the
low-rank update \(\Delta W\  = \ A\ B\ C\) contributes no additional
transformation at the start of training, allowing the model to preserve
its pre-trained behavior. The scaling parameter \(\alpha\) is
initialized to 1.0, allowing a balanced initial contribution of the
low-rank update. This initialization ensures that TLoRA begins training
from a stable configuration, effectively leveraging the representational
capacity of \(A\) and \(C\) while dynamically learning task-specific
transformations in \(B\).

\section{Empirical Experiments}

In this study, we evaluate the effectiveness of TLoRA on several
downstream tasks, comparing it against LoRA in terms of performance and
parameter efficiency. Can TLoRA perform competitively with LoRA, despite
having a fewer parameters? We evaluate our approach on four
classification tasks selected from the GLUE benchmark \cite{wang_glue_2018}:
MRPC (Microsoft Research Paraphrase Corpus), RTE (Recognizing Textual
Entailment), QNLI (Question-answering NLI), and SST-2 (Stanford
Sentiment Treebank). These tasks span a range of natural language
understanding challenges, from sentence similarity to sentiment
analysis, offering a comprehensive evaluation of the
models\textquotesingle{} capabilities.

We conduct our experiments using the pre-trained transformer model 
RoBERTa-large, a bidirectional encoder model designed for robust
language understanding tasks. Specifically, we leverage the MNLI
checkpoint \cite{williams2018broad,hu_lora_2021}, which has been fine-tuned on the Multi-Genre Natural
Language Inference (MNLI) dataset. This choice aligns with prior work, allowing for a direct
comparison of our proposed TLoRA method with existing adaptation
techniques under consistent conditions. The MNLI-tuned RoBERTa-large
provides a strong baseline, as it is optimized for handling complex
sentence-pair classification tasks, making it well-suited for evaluating
the effectiveness of low-rank adaptation methods like TLoRA. Similar to LoRA, we specifically target the linear layers within the attention submodules—the self.query and self.value projection layers, since these play a central role in the model’s ability to capture semantic relationships between tokens.

The objective of our experiments is to evaluate how well TLoRA performs on classification tasks in terms of both accuracy and computational efficiency, and how it compares against other adaptation methods such as LoRA. We
measure the impact of these techniques on the GLUE tasks by comparing
the models\textquotesingle{} accuracy across all tasks, as well as the
number of parameters that need to be fine-tuned.

\begin{table}[ht]
  \small
  \centering
  \caption{TLoRA training parameters and setup for the GLUE tasks}
  \label{tab:setup}
  \begin{tabular}{lcccc}
    \toprule
       & SST-2 & QNLI & RTE & MRPC \\
    \midrule
    Batch size    & 32&32&32&32\\
    Learning rate & 1e-3&1e-3&1e-3&1e-3\\
    Epochs        & 30&30&30&30\\
    Rank          & 32&32&32&32\\
    Optimizer     & AdamW&AdamW&AdamW&AdamW\\
    LR schedule   & Linear&Linear&Linear&Linear\\
    \bottomrule
  \end{tabular}
\end{table}

Our experimental setup, summarized in Table 2, includes fine-tuning
RoBERTa-large on four benchmark datasets: SST-2, QNLI, RTE, and MRPC.
For each dataset, we maintain consistent hyperparameters to ensure fair
comparisons across tasks. We use a batch size of 32 and fine-tune for 30
epochs per task. The low-rank adaptation rank is fixed at 32 for all
experiments to balance adaptation expressiveness with parameter
efficiency. Optimization is performed using the AdamW optimizer, with a
linear learning rate schedule applied to adjust the learning rate over
training. We also apply a dropout rate of 50\% to the low-rank update
\(\Delta h\) to prevent overfitting and improve generalization. This
unified setup allows us to systematically evaluate the effectiveness of
TLoRA in low-rank fine-tuning across diverse NLP tasks.

\begin{figure}[h]
  \centering
  \includegraphics[width=0.5\linewidth]{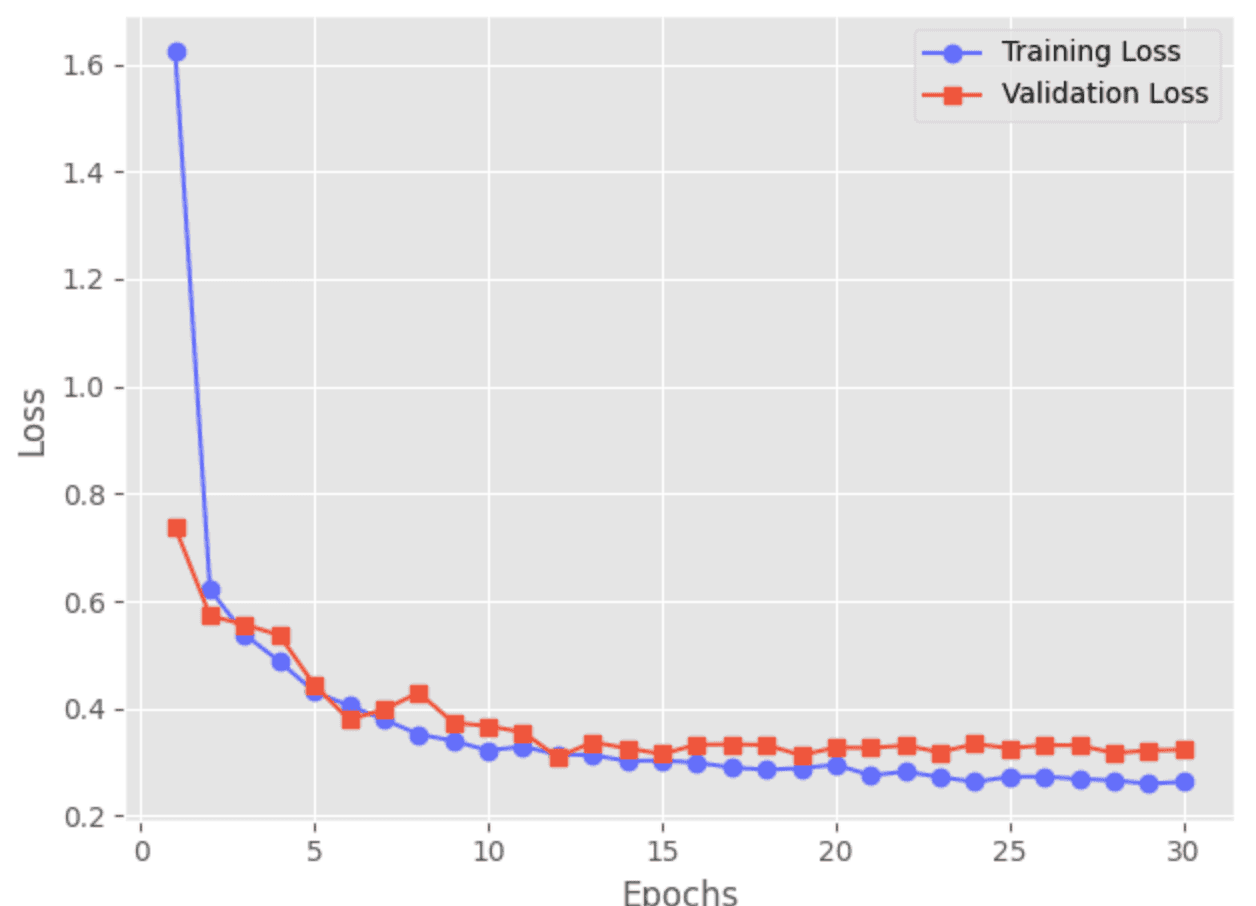}
  \caption{Training and validation loss curves for the MRPC
dataset. The figure demonstrates the stability of TLoRA during training
over 30 epochs.}
  \label{fig:3}
\end{figure}

\section{Results}
First, we present the training and validation loss curves for TLoRA, shown in
Figure 3. The figure is constructed for the MRPC dataset over 30 epochs.
TLoRA demonstrates stable training dynamics, with both training and
validation losses decreasing consistently throughout the epochs before
reaching a plateau. This stability highlights TLoRA's ability to
maintain effective learning with low-rank parameterization, avoiding
issues such as overfitting or loss divergence. The close alignment
between training and validation loss curves further illustrates
TLoRA\textquotesingle s generalization capability, suggesting that our
tri-matrix decomposition and adaptive scaling techniques successfully
capture task-relevant patterns without excessive parameter overhead.

\begin{table}[h]
  \small
  \centering

  \caption{Results for different adaptation methods on the GLUE benchmark. Reported metrics are accuracy (\%). The TLoRA results are based on our implementation, while the results for other methods are sourced from prior work \cite{hu_lora_2021,kopiczko_vera_2023}.}
  \label{tab:results_glue}
  \begin{tabularx}{\linewidth}{
      >{\raggedright\arraybackslash}X
      >{\centering\arraybackslash}X
      >{\centering\arraybackslash}X
      >{\centering\arraybackslash}X
      >{\centering\arraybackslash}X
      >{\centering\arraybackslash}X
      >{\centering\arraybackslash}X
      >{\centering\arraybackslash}X
    }
    \toprule
    \thead{Model}
      & \thead{Fine \\ Tuning\\Method}
      & \thead{Trainable\\Params}
      & \thead{SST-2}
      & \thead{QNLI}
      & \thead{RTE}
      & \thead{MRPC}
      & \thead{Avg} \\
    \midrule
    RoBERTa  & AdptP    & 3M       & 96.1 & 94.8 & 83.8 & 90.2 & 91.2 \\
             & AdptP    & 0.8M     & 96.6 & 94.8 & 80.1 & 89.7 & 90.3 \\
             & AdptH    & 6M       & 96.2 & 94.7 & 83.4 & 88.7 & 90.8 \\
             & AdptH    & 0.8M     & 96.3 & 94.7 & 72.9 & 87.7 & 87.9 \\
             & LoRA-FA  & 3.7M     & 96.0 & 94.4 & 86.1 & 90.0 & 91.6 \\
             & LoRA     & 0.8M     & 96.2 & 94.8 & 85.2 & 90.2 & 91.6 \\
             & VeRA     & 0.061M   & 96.1 & 94.4 & 85.9 & 90.9 & 91.8 \\
             & TLoRA    & 0.049M   & 95.3 & 92.1 & 87.5 & 89.3 & 91.0 \\
    \bottomrule
  \end{tabularx}
\end{table}

In our experiments, TLoRA demonstrates competitive performance across
multiple datasets while maintaining an exceptionally low parameter
footprint. As shown in Table 3, TLoRA achieves an average accuracy of
91.0\% across the SST-2, QNLI, RTE, and MRPC benchmarks with only
0.049M trainable parameters---significantly fewer than methods such as
Adapter (AdptP) and LoRA, which use parameter counts ranging from 0.8M
to 6M. Notably, TLoRA achieves a high 87.5\% accuracy on the RTE task,
outperforming larger configurations like AdptH and LoRA-FA methods. This parameter
efficiency can be attributed to TLoRA's tri-matrix decomposition and
adaptive scaling, which allow the model to capture complex task-specific
information with minimal trainable weights. Although TLoRA's accuracy on
certain datasets, such as QNLI, is slightly lower than other methods,
its trade-off in parameter efficiency and competitive accuracy across
tasks showcases its potential as a highly scalable, efficient adaptation
technique for large language models.

\section{TLoRA Adaptation Dynamics}
In this section, we provide an in-depth analysis of the behavior and
adaptation dynamics of TLoRA; and compare its properties with those of
LoRA to better understand its low-rank adaptation capabilities. For
demonstration, we analyze the TLoRA-trained model on the MRPC dataset.

\begin{figure}[h]
  \centering
  \includegraphics[width=0.8\linewidth]{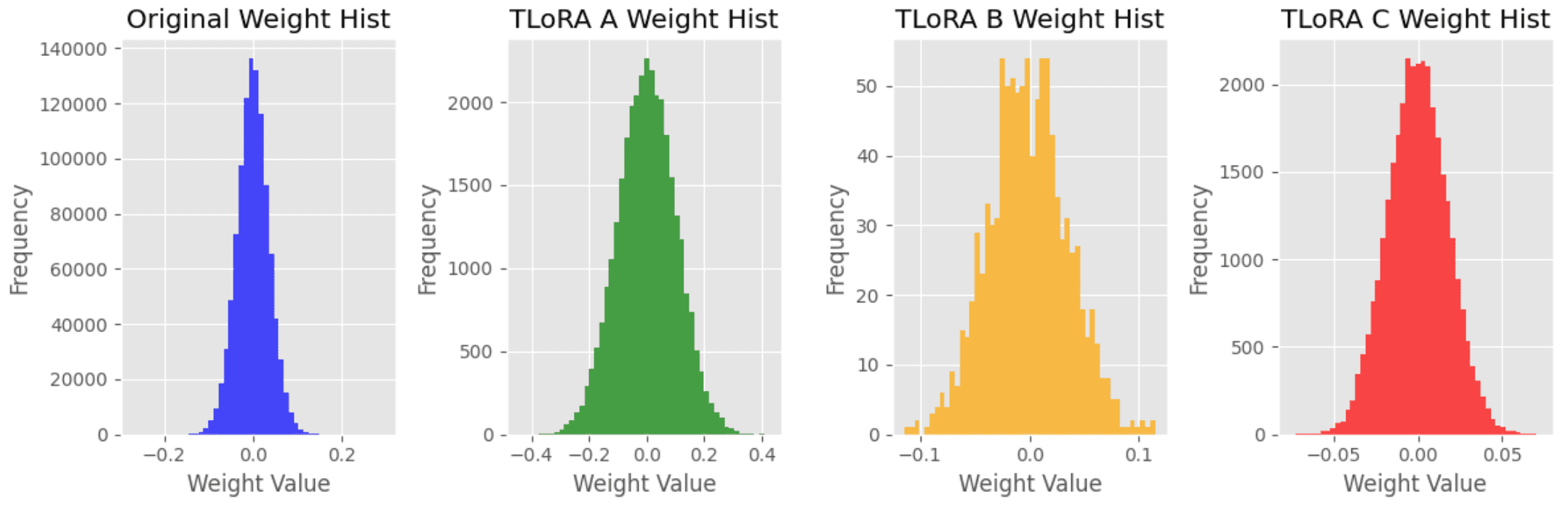}
  \caption{Weight distribution histograms for the original weight matrix and
the TLoRA matrices \(A\), \(B\), and \(C\). The matrices \(A\) and \(C\)
are randomly initialized and remain fixed, following a Gaussian
distribution. In contrast, the trainable matrix \(B\), which is
initialized to zero, evolves during training and adopts a Gaussian-like
distribution, highlighting the effectiveness of TLoRA\textquotesingle s
tri-matrix decomposition.}
  \label{fig:4}
\end{figure}

We first examine the weight distributions of TLoRA's tri-matrix
components, as depicted in Figure 4. This is constructed for layer 0 and
the query (\(q\)) matrix. We find clear evidence that the adaptation
aligns well with natural statistical properties and maintains stability
across updates. Matrices \(A\) and \(C\) are randomly initialized using
a Gaussian distribution to provide a stable, structured space for the
low-rank adaptation. This initialization allows \(A\) and \(C\) to
effectively capture diverse input-output interactions while remaining
fixed throughout training. In contrast, matrix \(B\), which is
initialized to zeros, evolves significantly over training.
Post-training, the weight distribution of \(B\) closely resembles a
Gaussian pattern, indicating that it has learned a meaningful structure
aligned with task-specific representations.

\begin{figure}[h]
  \centering
  \includegraphics[width=0.7\linewidth]{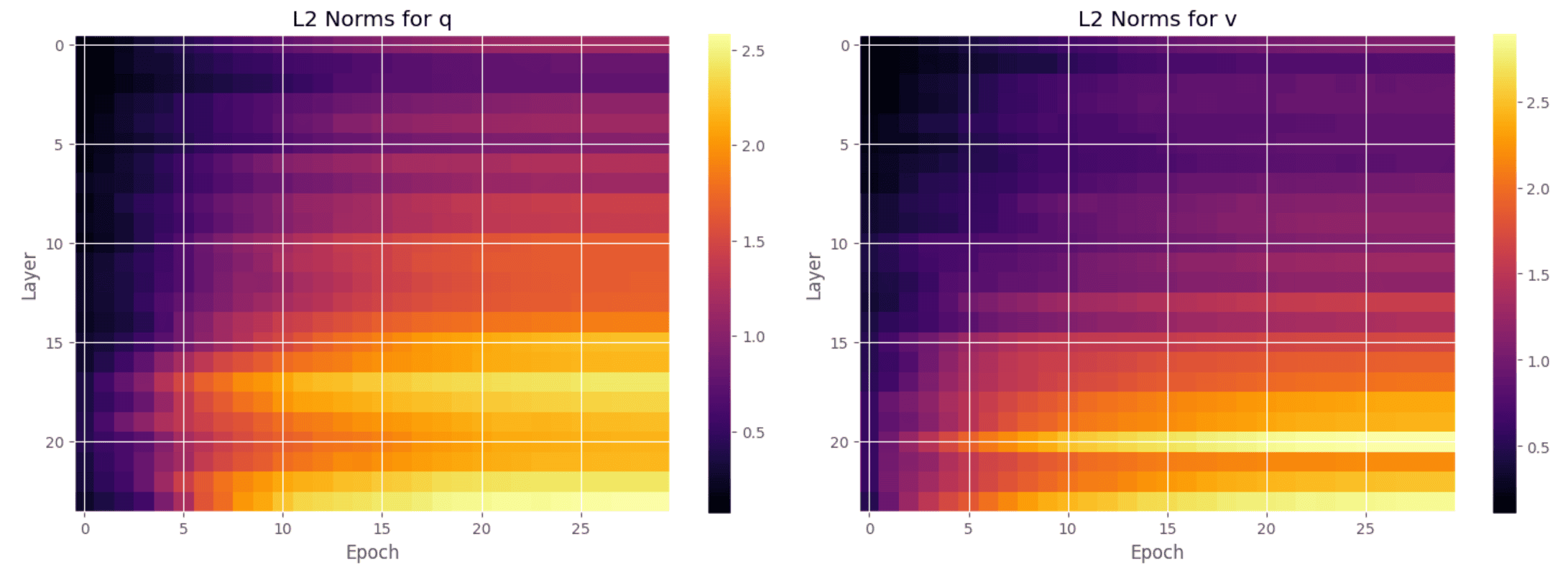}
  \caption{Evolution of L2 norms for the TLoRA \(B\) matrix over training
epochs.}
  \label{fig:5}
\end{figure}

\subsection{TLoRA parameter behavior}

The evolution of TLoRA\textquotesingle s L2 norms across training epochs
provides further insights into the layer-wise adaptation dynamics, as
shown in Figure 5. At the start of training (epoch 1), the L2 norms of
TLoRA parameters for both the query and value matrices are close to
zero, aligning with the initial zero-initialization of matrix \(B\) in
TLoRA. This zero starting point ensures that TLoRA begins with minimal
influence on the pre-trained model's output, allowing for a smooth and
gradual adaptation to the target task as training progresses.

As training advances, we observe that the L2 norms increase, but the
growth rates vary significantly across layers. This variation suggests
that each layer's query and value projections adapt differently, likely
reflecting the differential relevance of specific layers to the given
task. The increase in L2 norm values with training indicates that TLoRA
is progressively capturing task-specific information, with each
layer\textquotesingle s TLoRA parameters contributing to a learned
adaptation in the low-rank subspace. This dynamic growth in the L2 norms supports our hypothesis that TLoRA
efficiently leverages the minimal trainable matrix~\(B\) to encode
essential adaptations without excessive parameter overhead. The pattern observed in the L2 norms
confirms that TLoRA is effectively engaging in layer-wise task-specific
learning, adapting the low-rank representations as needed for each layer
while maintaining efficient parameter utilization. This controlled and
progressive adaptation process underscores TLoRA\textquotesingle s
capability to balance parameter sparsity with effective learning, a key
objective in the design of our tri-matrix structure.

\begin{figure}[ht]
  \centering
  \includegraphics[width=0.7\linewidth]{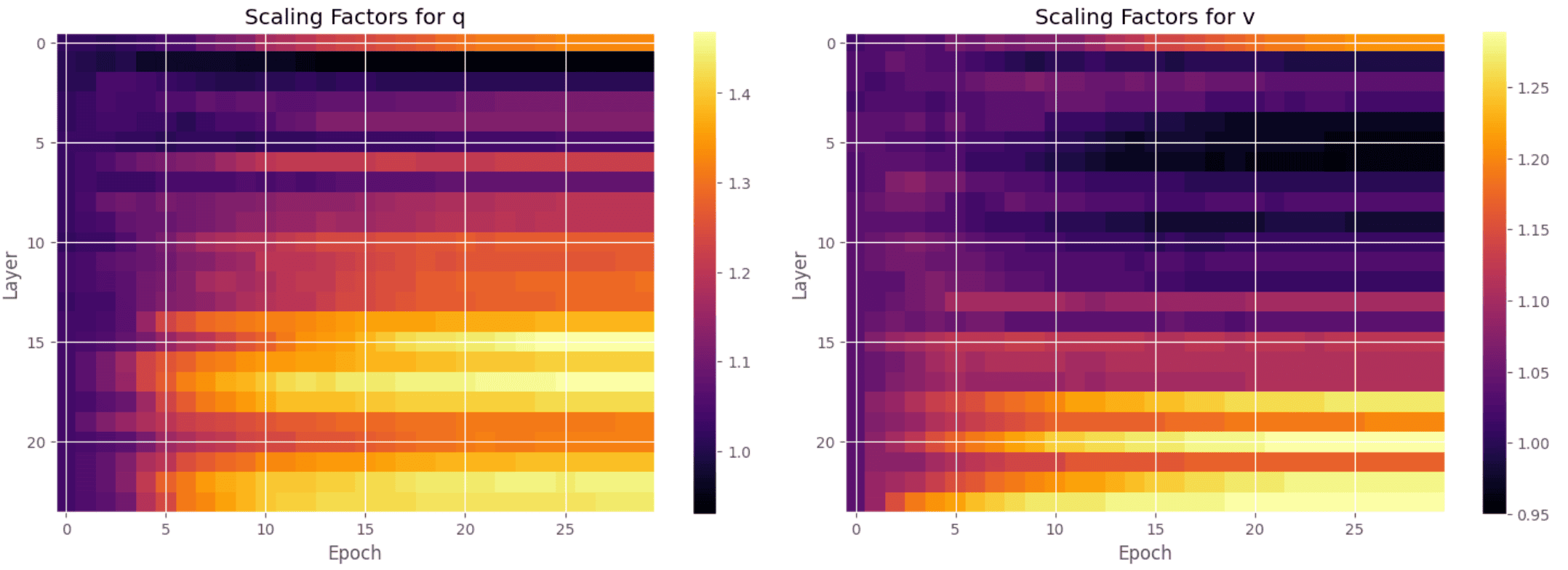}
  \caption{Evolution of scaling factors over training epochs for TLoRA. The
scaling factors, which are learnable and layer-specific, start from an
initial value of 1 and exhibit significant variability across layers.}
  \label{fig:6}
\end{figure}

In examining the evolution of TLoRA's learned scaling factors over
epochs, we observe substantial layer-wise variability, as depicted in
Figure 6. Initially, each scaling factor starts at 1, providing a
uniform impact across layers. However, as training progresses, we see
distinct divergence in these factors: some layers show a significant
increase in scaling values, while others decrease, reflecting a dynamic,
adaptive adjustment based on each layer\textquotesingle s contribution
to task performance. Furthermore, within each layer, scaling factors
differ between the query (\(q\)) and value (\(v\)) matrices, suggesting
that TLoRA is refining its adaptation granularity to capture the unique
functional roles of \(q\) and \(v\) within each layer.

This learned variability in scaling factors is a crucial component of
TLoRA's adaptation mechanism. By allowing scaling factors to adjust
independently across layers and between \(q\) and \(v\), TLoRA
dynamically modulates the influence of low-rank updates, enhancing the
model\textquotesingle s ability to capture task-specific nuances without
disrupting pre-trained knowledge. Higher scaling factors in certain
layers, for instance, indicate that those layers require more pronounced
adaptations, which TLoRA accommodates by increasing the influence of its
low-rank component in those areas. Conversely, reduced scaling factors
in other layers suggest that minimal alteration is necessary, enabling
TLoRA to selectively constrain changes where the pre-trained parameters
are already well-aligned with the task.

This layer- and component-wise adaptability confirms TLoRA's capacity
for flexible, fine-grained adjustment, allowing the model to engage in
task-specific tuning without excessive parameter growth. The learned
scaling factors thus act as a refined control mechanism, optimizing the
integration of TLoRA\textquotesingle s low-rank components and
contributing to its effectiveness as an efficient yet expressive
fine-tuning approach.

\begin{figure}[h]
  \centering
  \includegraphics[width=0.5\linewidth]{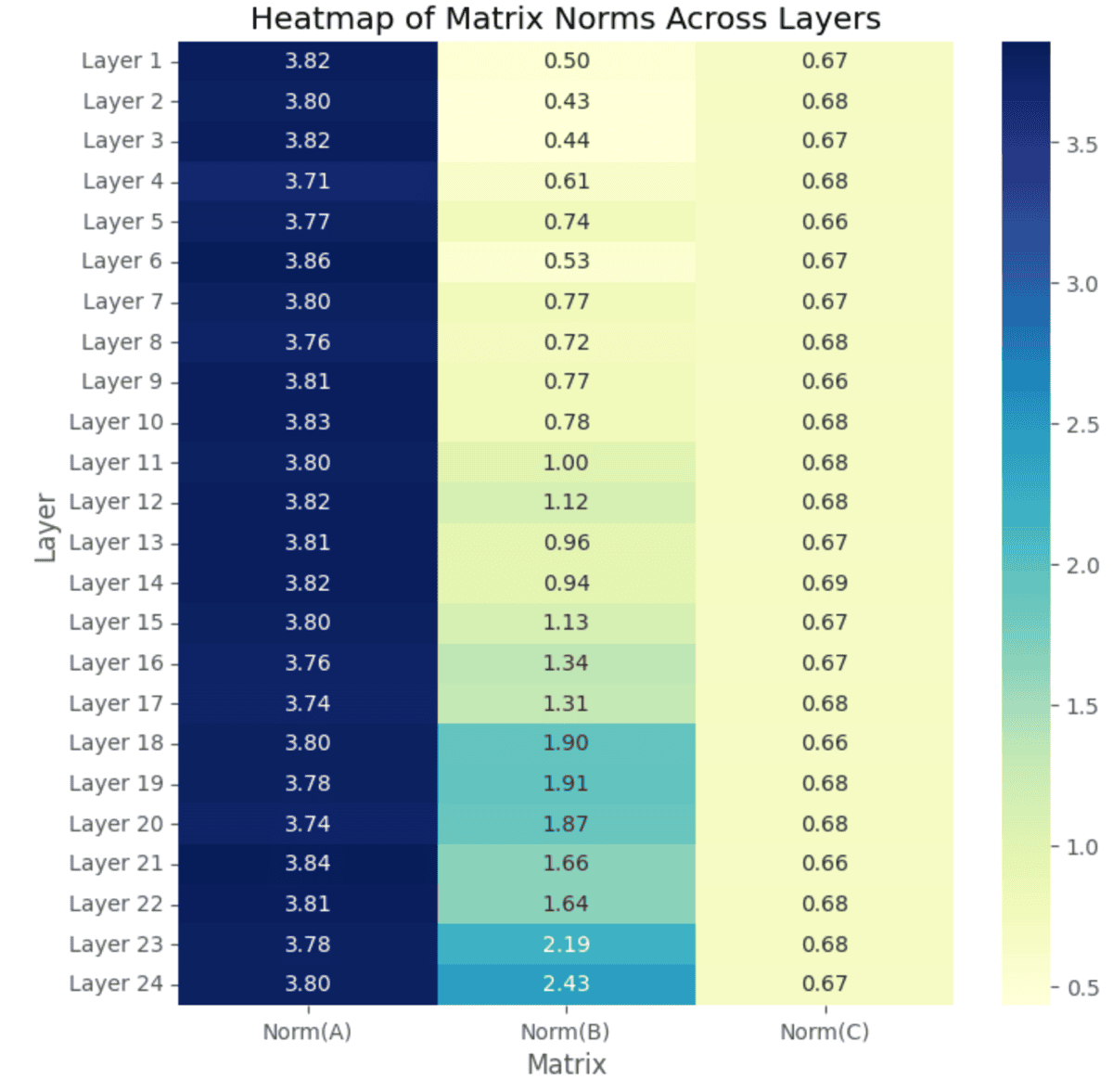}
  \caption{Heatmap of layer normalization values for the query
(\(q\)) matrices across the tri-matrix components \(A,\ B,\ C\). The
fixed matrices \(A\) and \(C\) show no variability in their layer
normalization values, as expected. In contrast, the trainable matrix \(B\)
exhibits significant variability across layers, which corresponds to the
learned task-specific adaptations.}
  \label{fig:7}
\end{figure}

Figure 7 presents a heatmap of the layer normalization values (matrix L2 norms) for the
query (\(q\)) matrices across the tri-matrix components \(A,\ B,\ C\).
As expected, since matrices \(A\) and \(C\) are randomly initialized and
fixed throughout training, we observe minimal to no variability in their
layer normalization values, yielding uniform heatmap patterns. This
stability reflects that \(A\) and \(C\) retain their initialized
structure and act as static projections.

In contrast, matrix \(B\), which is trainable, exhibits clear
variability across layers in its layer normalization values. This
variability indicates that \(B\) adapts dynamically during training,
with each layer capturing distinct task-specific transformations.
Interestingly, the layers with higher scaling factors, as previously
discussed, also show correspondingly higher layer normalization values
for \(B\). This relationship suggests that the layers requiring larger
adaptations to align with task demands exhibit more pronounced changes
in \(B\), which are reflected both in their scaling factors and in their
layer normalization profiles.

These findings reinforce the role of \(B\) as the adaptable core of
TLoRA's low-rank adaptation framework, responding layer by layer to
task-specific requirements while leveraging the stable structures
provided by \(A\) and \(C\). The alignment between scaling factors and
layer normalization values further highlights TLoRA's ability to target
specific layers for adaptation intensity, enhancing both its flexibility
and efficiency in fine-tuning transformer models.

\subsection{TLoRA resembles LoRA}

To investigate the similarity between TLoRA and LoRA, we analyze whether
TLoRA follows a comparable update trajectory to that of LoRA during
training. Using the same experimental procedure as outlined in the LoRA
paper, we first train a LoRA model. We subsequently evaluate the
adaptation dynamics of both LoRA and TLoRA methods on the GLUE MRPC
dataset, providing a detailed comparison of their behavior and
performance during fine-tuning.

\begin{figure}[ht]
  \centering
  \includegraphics[width=0.6\linewidth]{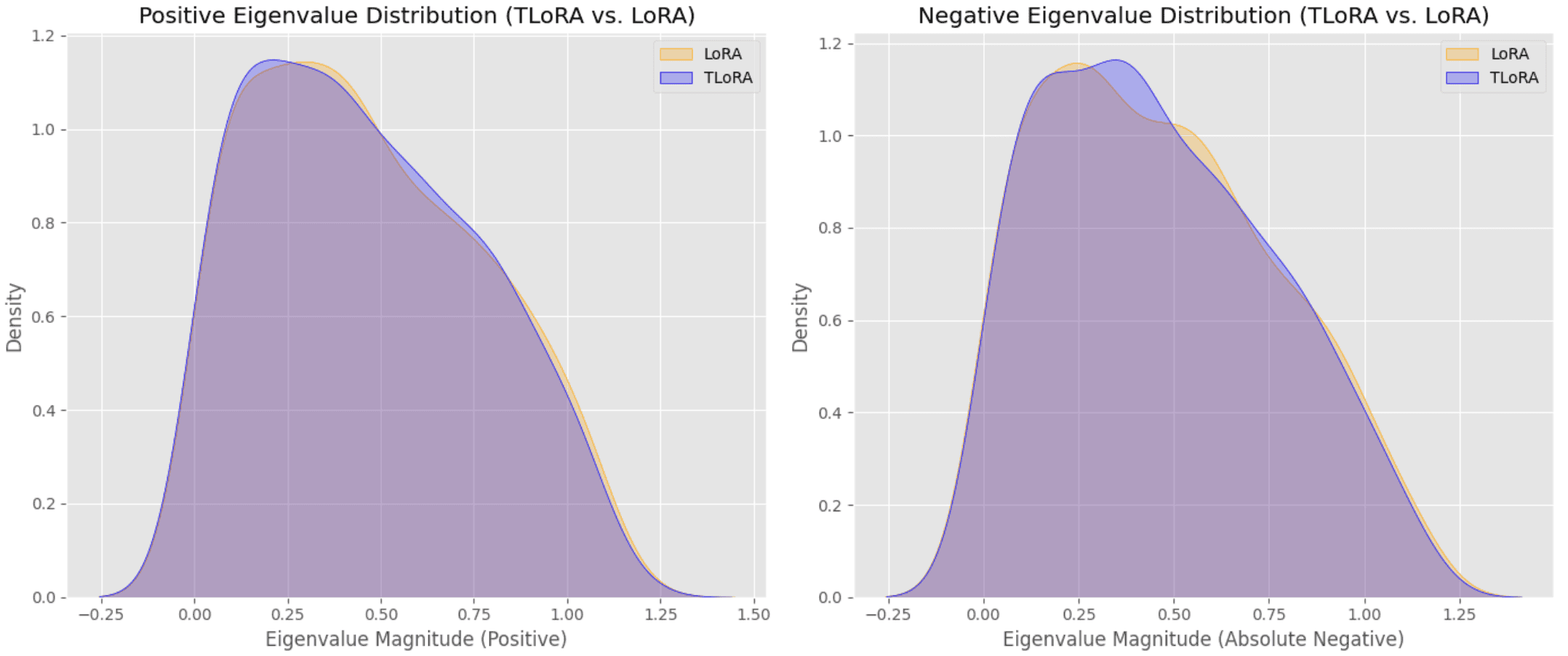}
  \caption{Eigenvalue distributions of the learned parameter updates for TLoRA
and LoRA. The figure compares the eigenvalue distributions of the
updates for both methods, showing that the distributions closely align
in terms of both positive and negative eigenvalues.}
  \label{fig:8}
\end{figure}

In Figure 8, we present the eigenvalue distributions of the learned
parameter updates for both LoRA and TLoRA. Here, we plot positive and negative eigenvalues separately (magnitudes shown). The
distributions align closely between the two methods, with TLoRA's
eigenvalues closely mirroring the spread and magnitude of LoRA's. This
similarity in eigenvalue behavior suggests that, despite
TLoRA\textquotesingle s use of a tri-matrix decomposition and fewer
trainable parameters, it is able to capture the essential directions of
adaptation in a manner analogous to LoRA. The resemblance in eigenvalue
distributions underscores that TLoRA achieves an effective and efficient
approximation of LoRA's adaptation pathway. This observation supports
the hypothesis that TLoRA's tri-matrix design, even with fixed
parameters in \(A\) and \(C\), can reach a comparable low-rank solution,
capturing the core task-specific transformations in a similarly
structured manner.

\begin{figure}[h]
  \centering
  \includegraphics[width=0.5\linewidth]{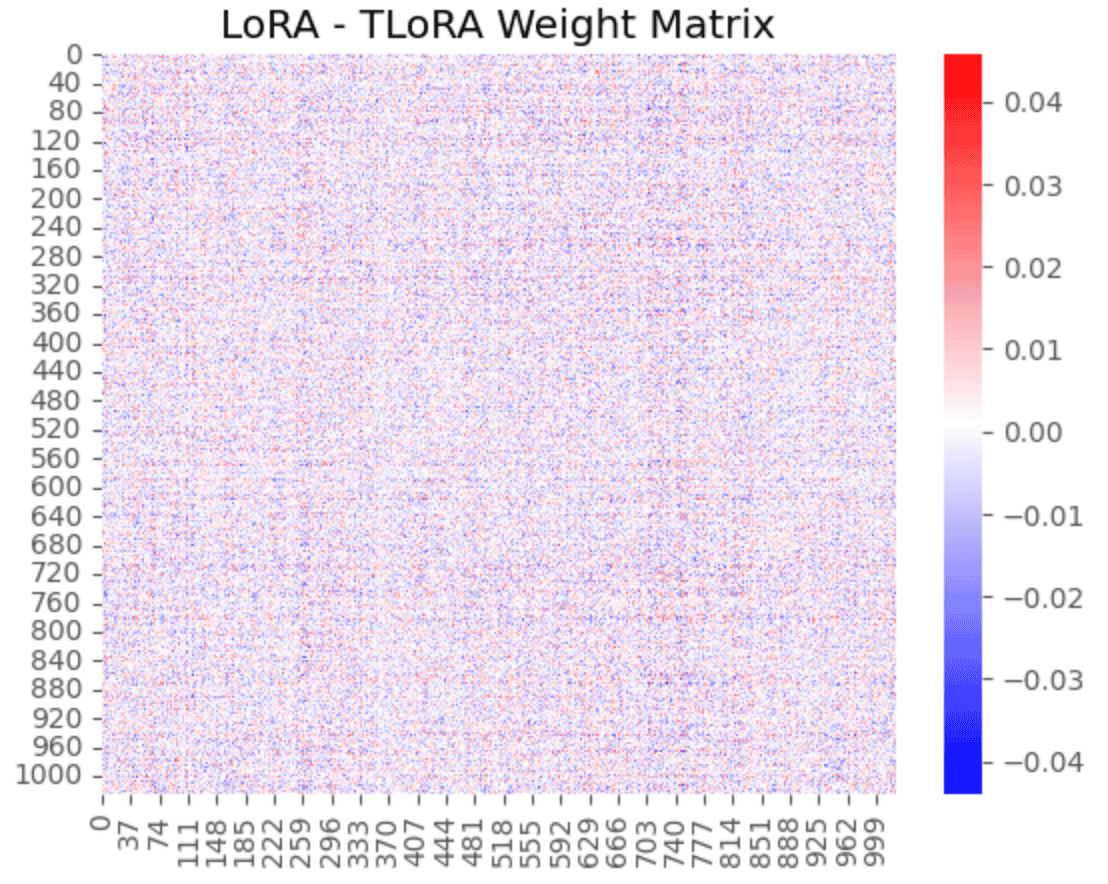}
  \caption{The figure shows the element-wise difference of the
weight matrices between LoRA and TLoRA for a specific layer. The
differences are close to zero, indicating that the adaptation behavior
of TLoRA closely resembles that of LoRA.}
  \label{fig:9}
\end{figure}

\begin{figure}[h]
  \centering
  \includegraphics[width=0.5\linewidth]{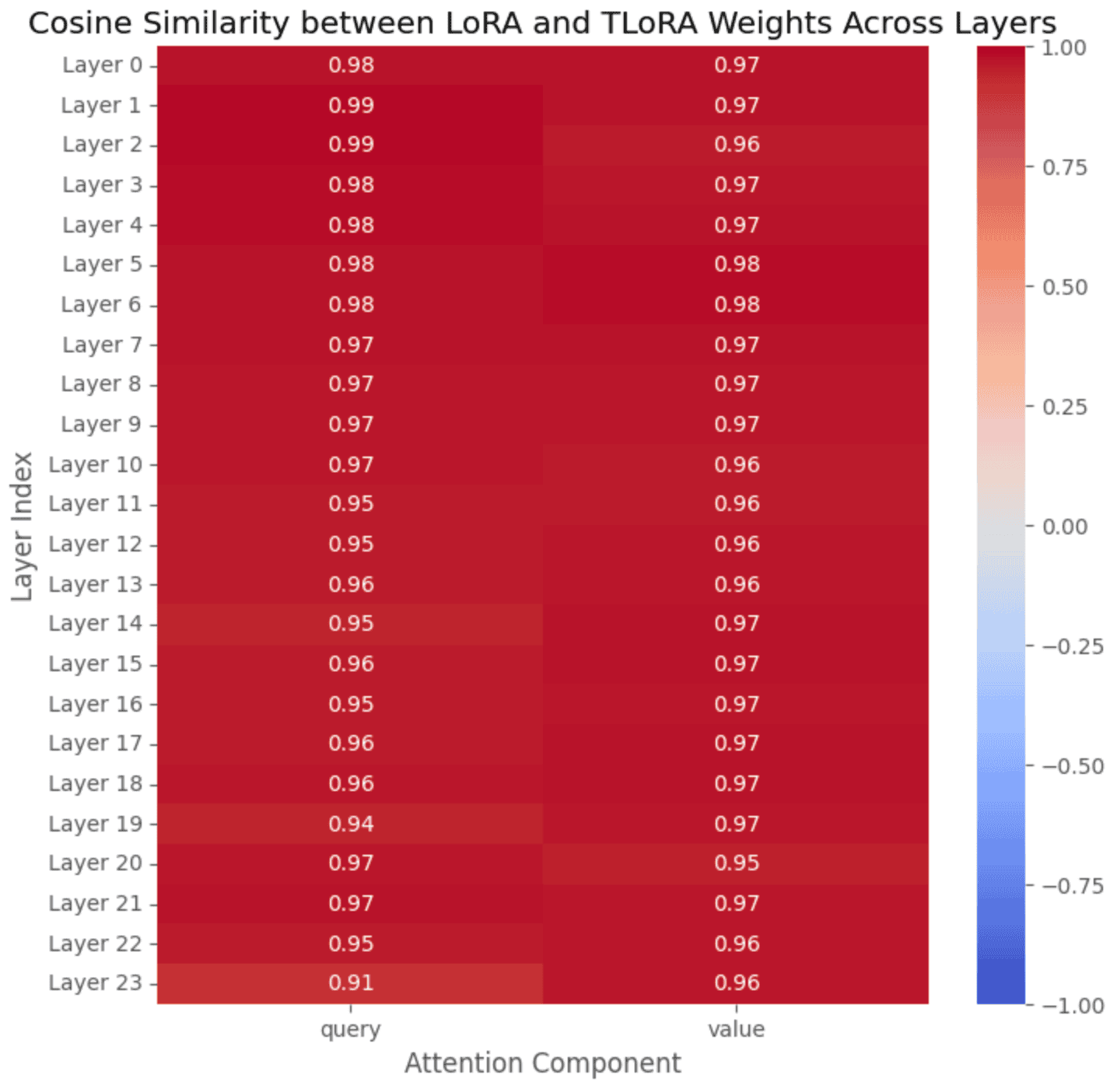}
  \caption{Cosine similarity values across all layers for the query (q) and
value (v) attention components between TLoRA and LoRA.}
  \label{fig:10}
\end{figure}

This resemblance between TLoRA and LoRA is further highlighted by
examining the element-wise differences between their adaptation
matrices. In Figure 9, we visualize the difference between the LoRA and
TLoRA matrices for the query (\(q\)) component in the first layer (layer
0). As shown, the values in the LoRA--TLoRA matrix are close to zero
across most elements, indicating minimal divergence between the two
methods in their parameter updates at this layer.

The near-zero differences suggest that, despite structural distinctions
and TLoRA's added scaling and fixed matrices \(A\) and \(C\), both
methods arrive at similar low-rank parameter adjustments for this layer.
This outcome further supports the idea that TLoRA effectively mirrors
LoRA's adaptation without requiring identical parameter configurations,
demonstrating that TLoRA can approximate LoRA's solution while
leveraging its tri-matrix design.

To further quantify the similarity between LoRA and TLoRA, we calculate
the cosine similarity between their learned parameter updates across all
layers for both the query (\(q\)) and value (\(v\)) attention
components. As shown in Figure 10, the cosine similarity values are
consistently close to 1 across layers, indicating a high degree of
alignment in the adaptation directions achieved by TLoRA compared to
LoRA. This high similarity suggests that, despite TLoRA's unique
tri-matrix decomposition with two fixed matrices and a learnable scaling
factor, it effectively approximates the directional adjustments made by
LoRA. The close alignment for both (\(q\)) and (\(v\)) components
indicates that TLoRA captures the essential transformations required for
task-specific adaptation in a manner nearly indistinguishable from LoRA.

\section{Conclusion}
Building on the foundational work of Low-Rank Adaptation (LoRA), we
introduce TLoRA, a novel fine-tuning technique designed to enhance model
adaptability and performance with computational efficiency. TLoRA
introduces a tri-matrix decomposition to adapt pre-trained language
models, utilizing the matrices \(A,\ B,\ C\) to compute a low-rank
update to the model\textquotesingle s weights. The contribution of this
update is controlled by a trainable scaling factor \(\alpha\), which is
learned during training. This approach strikes a balance between
computational efficiency and adaptation flexibility, enabling effective
model fine-tuning with a minimal increase in parameters. By using
non-trainable matrices \(A\) and \(C\) and allowing \(B\) to be
trainable, TLoRA enables efficient adaptation while maintaining the core
functionality of the original pre-trained model.

TLoRA offers a robust and efficient method for fine-tuning large
language models, pushing the boundaries of model adaptability and
performance. This novel approach showcases the potential for extending
low-rank adaptation techniques to achieve greater efficiency and
effectiveness in the evolving landscape of large language models.

\bibliographystyle{splncs04}
\bibliography{references.bib}

\end{document}